\pdfoutput=1
\documentclass[letterpaper, 10 pt, conference]{ieeeconf}  

\IEEEoverridecommandlockouts                              
\usepackage{cite}
\usepackage[pdftex]{graphicx}
\graphicspath{{../figures/}}	
\usepackage{psfrag}			
\usepackage{amsmath}		
\usepackage{cancel}
\usepackage{amsfonts}		
\usepackage{url}				
\usepackage{here}				
\usepackage{color}			
\usepackage{balance}    
\usepackage{float}
\usepackage{bm}
\usepackage{bbm}
\usepackage{amsmath,amsfonts,amssymb}
\usepackage[]{algorithm2e}
\usepackage[utf8]{inputenc}
\usepackage{verbatim}
\usepackage{scalerel}
\usepackage{amssymb}
\usepackage{subfiles}
\usepackage{leftidx}
\usepackage[flushleft]{threeparttable}
\usepackage{chngcntr}

\newcommand{\Bezier}{B\'ezier}
\newtheorem{assumption}{\textbf{Assumption}}

\title{\LARGE \bf Kinodynamic Motion Planning for Multi-Legged Robot Jumping via Mixed-Integer Convex Program}
\author{Yanran~Ding$^1$,
	Chuanzheng~Li$^1$,
	and Hae-Won Park$^2$
\thanks{This project is supported in part by NAVER LABS Corp. under grant 087387, Air Force Office of Scientific Research under grant FA2386-17-1-4665, and National Science Foundation under grant 1752262.}
\thanks{Yanran Ding and Chuanzheng Li are with the $^1$ Department of Mechanical Science and Engineering, University of Illinois at Urbana-Champaign, Urbana, IL, 61820 USA. (email:\{yding35,cli67\}@illinois.edu)}
\thanks{Hae-Won Park is with the $^2$ Department of Mechanical Engineering, Korea Advanced Institute of Science and Technology, Daejeon-34141, South Korea. (email:haewonpark@kaist.ac.kr)}
}

\begin{document}
	\maketitle
	\thispagestyle{empty}
	\pagestyle{empty}
	
	\begin{abstract}
		This paper proposes a kinodynamic motion planning framework for multi-legged robot jumping based on the mixed-integer convex program (MICP), which simultaneously reasons about centroidal motion, contact points, wrench, and gait sequences. This method uniquely combines configuration space discretization and the construction of feasible wrench polytope (FWP) to encode kinematic constraints, actuator limit, friction cone constraint, and gait sequencing into a single MICP. The MICP could be efficiently solved to the global optimum by off-the-shelf numerical solvers and provide highly dynamic jumping motions without requiring initial guesses. Simulation and experimental results demonstrate that the proposed method could find novel and dexterous maneuvers that are directly deployable on the two-legged robot platform to traverse through challenging terrains.
	\end{abstract}

	\section{Introduction}
	\label{sec:introduction}
	The ability to perform dynamic motions such as leaping over gaps and jumping on high platforms is a unique advantage of legged systems. Coordinating multiple limbs to execute dynamic motions is a challenging problem since it involves both continuous and discrete variables. This problem requires decision making in a semi-continuous search space, which involves continuous variables describing robot state, contact positions, and contact wrenches; It also involves discrete variables such as the gait sequence.
	
	Many methods have been developed to solve this problem. For example, the trajectory optimization (TO) approach locally improves upon an initial motion plan by solving a general nonlinear optimization problem using a gradient-based nonlinear solver. There has been tremendous progress in using TO to solve locomotion problems. MIT Cheetah 2 robot could jump over obstacles by solving nonlinear constrained optimization online \cite{haewon_IJRR}. Optimized jumping trajectories are generated offline \cite{Quan_ICRA} and implemented on MIT Cheetah 3 \cite{MIT_Cheetah3}. Linear complementary problems (LCP) are formulated in \cite{Posa_13_IJRR} to generate trajectories without \textit{a priori} contact scheduling. Dynamic movements without scheduled contact are also generated in \cite{Mastalli16} using a hierarchical framework. The combined planning problem is solved in \cite{mordatch2012discovery} by incorporating all constraints into the objective function, and solve unconstrained nonlinear programming (NLP). Legged locomotions with gait sequences are generated on non-flat terrain in \cite{Wrinkler18} in a single TO formulation using a phase-based parameterization method. These methods either rely on explicit contact schedules or require solving a large NLP. The size and non-convexity of these problems imply that the nonlinear solver is only effective searching for a local minima around the initial guess. Hence, proper initialization of the optimization is crucial in finding a feasible solution. Besides, the infeasible status returned by a local NLP solver is not informative since one is not sure whether the planning problem itself is infeasible or it is not initialized properly.

	\begin{figure}
		\centering
		\resizebox{0.9\linewidth}{!}{\includegraphics{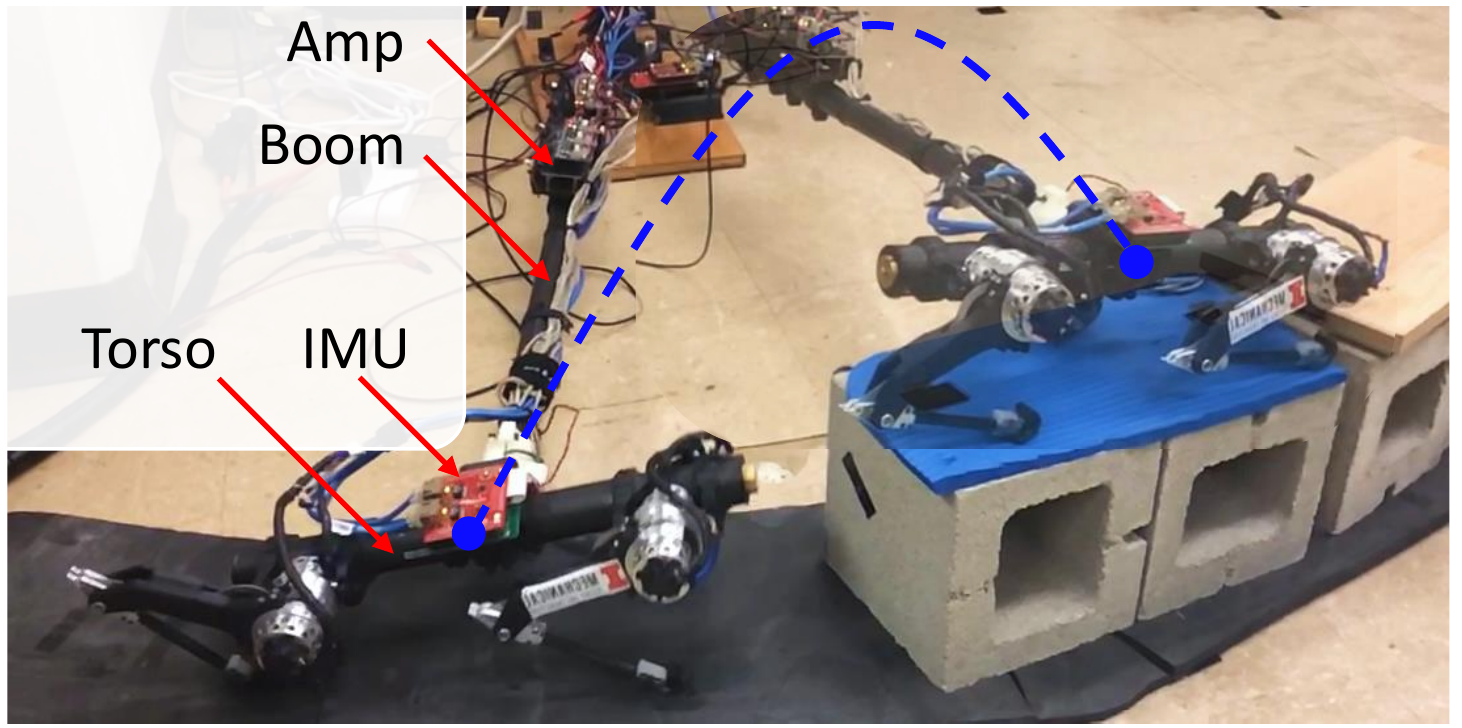}}
		\caption{The two-legged planar robot executing a dynamic jump to mount an obstacle 80\% of its height.}
		\label{fig:jumpUp}
	\end{figure}

	Mixed-integer convex optimization does not rely on the initial seed and warrants global solution \cite{boyd2004convex}. With the recent advancement in numerical solvers, a medium-sized mixed-integer convex programming (MICP) can be solved efficiently by off-the-shelf solvers such as gurobi \cite{gurobi}, Mosek \cite{mosek} and CPLEX \cite{CPLEX}. Due to its feature that warrants a global solution with either global optimality or infeasibility certificate, MICP has found many applications in robotics. For example, it has been used in global inverse kinematics \cite{Dai_IJRR}, grasping \cite{9120282}, footstep planning \cite{Deits_footstep}, quadruped locomotion planning \cite{BAC}, and aggressive legged locomotion \cite{valenzuela2016mixed}, \cite{Ding_18}.

	This paper presents a novel MICP-based kinodynamic motion planning framework for aggressive jumping motions while considering joint torque constraints. This method simultaneously plans centroidal motion, contact location, contact wrench, and gait sequence for a planner two-legged robot. To address the non-convex torque constraints due to the trigonometric terms in the Jacobian matrix, we adopt the notion of feasible wrench polytope (FWP) \cite{Romeo_18}. The joint torque constraint is approximated as a polytope containment problem over disjoint convex sets from the discretization of the configuration space (C-space). Similarly, bilinear terms are approximated using the McCormick Envelope \cite{mccormick1976computability} relaxation, hence resulting in a mixed-integer convex formulation. Our method reduces the number of decision variables by setting up a preferred ordering on the robot stance modes. As a result, any robot-state-trajectory in the C-space implies a unique gait sequence. The proposed method is evaluated on a hardware platform, as shown in Fig. \ref{fig:jumpUp}. The trajectory that guides the robot to jump on a platform is obtained without providing initial guesses to the MICP, which is not possible for local optimization based trajectory optimization methods.

	The paper is organized as follows: Section \ref{sec:technical_approach} introduces the single rigid body dynamics model, mixed-integer wrench constraint, and other constraints including aerial phase continuity and contact location choice. Section \ref{sec:results} presents the simulation and experiment results of the highly dynamic motion planning. Section \ref{sec:discusstion} discusses the results from this work and Section \ref{sec:conclusion} provides the concluding remarks.

	\section{Technical Approach}
	\label{sec:technical_approach}
	
	In this section, we briefly introduce the robot dynamic model before the construction of configuration space discretization and feasible wrench polytopes that are used in our formulation. And we also formalize our mixed-integer formulation with assumptions and constraints for aerial phase and foothold position choice.
	
	\subsection{Dynamic Model}\label{sec:dynamics}
	
	A simple model captures the major dynamical effect and reduce the number of optimization variables. As shown in Fig. \ref{fig:jumpUp}, the robot legs are made of light-weight material such as 3D printed parts and carbon fiber tubes, which results in the leg mass being less than 10\% of the total mass. Hence, we make the following assumption to simplify the dynamic model.
	\begin{assumption}[Light Legs]\label{as:light_leg}
		The leg mass is negligible.
	\end{assumption}
	
\begin{figure}
		\centering
		\resizebox{1\linewidth}{!}{\includegraphics{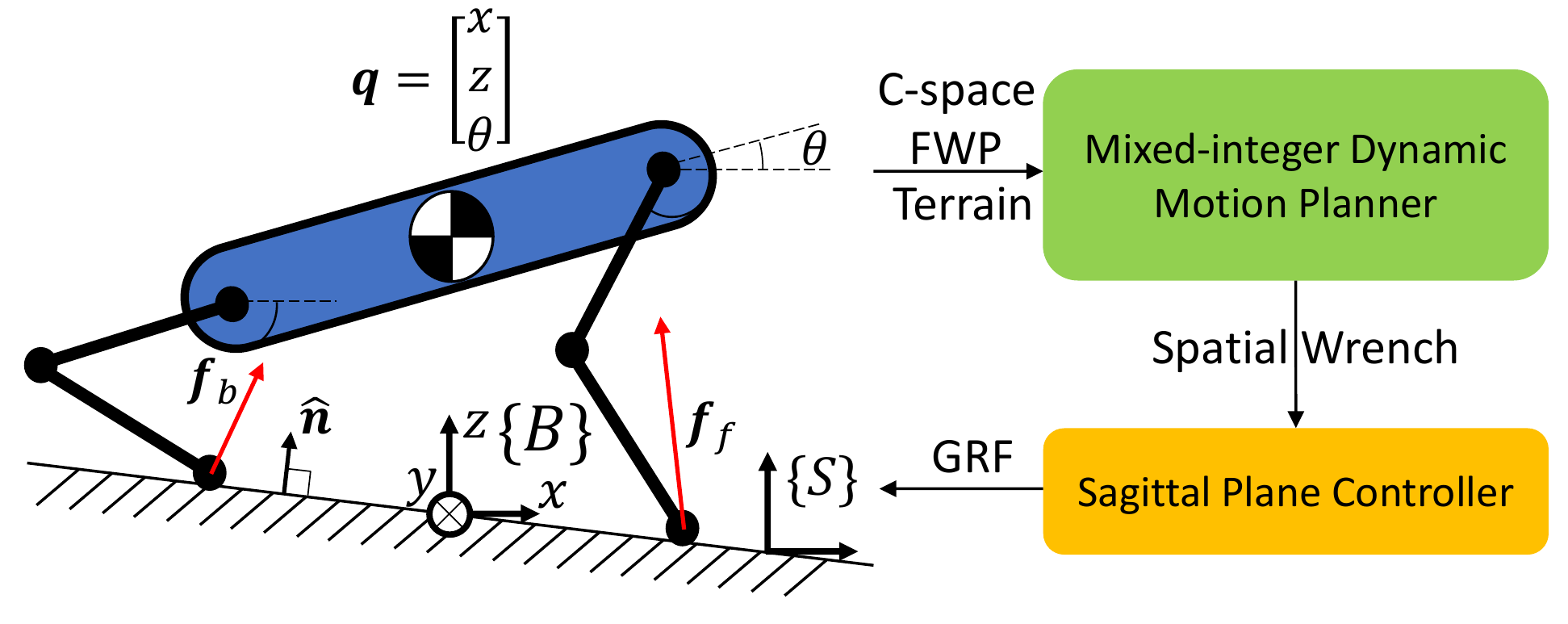}}
		\caption{Overview of the kinodynamic motion planning framework, which computes the motion and contact in a single MICP. The coordinate definition for $\{S\}$ (world frame) and $\{B\}$ (local frame) could be found on the left panel.}
		\label{fig:schematics}
	\end{figure}

	Based on assumption \ref{as:light_leg}, we employ the Centroidal Dynamics (CD) model \cite{orin2013centroidal}, specifically, the 2D single rigid body model which only considers the torso dynamics. As shown in Fig. \ref{fig:schematics}, the configuration of the robot could be represented by the Special Euclidean Group $SE(2)$ parameterized by $\bm{q} = [x,z,\theta]^T$, where $[x,z]^T$ is the location of the center of mass (CoM) and $\theta$ is the pitch angle. The input to the system is the spatial wrench $\bm{\mathcal{F}}_s$, and the dynamic model of the robot is
	\begin{equation}\label{eq:dyn}
	\ddot{\bm{q}} = \begin{bmatrix}
	\ddot{x}\\
	\ddot{z}\\
	\ddot{\theta}
	\end{bmatrix}=
	\bm{D}^{-1}\bm{\mathcal{F}}_s + \bm{a}_g,
	\end{equation}
	where $\bm{D}=\text{diag}(m,m,I_{\theta})$ is the inertia tensor; diag$(\cdot)$ creates diagonal matrices; $m$ is the total mass and $I_{\theta}$ is the moment of inertia around the CoM along the $z$-axis; $\bm{a}_g=[0,-g,0]^T$ is the gravitational acceleration vector. Assuming there are $N_c$ contact points, each with a ground reaction force (GRF) $\bm{f}_i\in\mathbb{R}^2$, the spatial wrench is given by 
	$\bm{\mathcal{F}}_s$
	\begin{equation}\label{eq:wrench}
	\bm{\mathcal{F}}_s = \sum_i^{N_c} \begin{bmatrix}
	\bm{f}_i\\
	\tau_i^y
	\end{bmatrix},
	\end{equation}
	where $\tau_i^y$ is the moment generated by $\bm{f}_i$, $\tau_i^y = \bm{r}_i \wedge \bm{f}_i$; $\bm{r}_i$ is the vector from CoM to the $i^{th}$ contact point; $\wedge:\mathbb{R}^2\times\mathbb{R}^2\rightarrow\mathbb{R}$ is the wedge product for two 2-dimensional vectors. The positive direction of the $y$-axis is pointing into the paper.
	
	Fig. \ref{fig:schematics} shows a schematics of the robot with frame definitions. The origin of the body-fixed frame $\{B\}$ is located at the middle point of the foot contact points. The axes of $\{B\}$ are aligned with that of the world frame $\{S\}$. The configuration of the robot is represented using two variables. The local configuration $\bm{q}$ is expressed in frame $\{B\}$; the global variable is defined in frame $\{S\}$, named the touchdown state $\bm{q}^{TD}=[x^{TD},z^{TD},\theta^{TD}]^T\in\mathbb{R}^3$, where $[x^{TD},z^{TD}]^T$ is the origin of frame $\{B\}$ and $\theta^{TD}$ is equal to the slope of the current terrain. This dichotomy of global and local states is convenient for imposing constraints of stance phase on the local state and choosing contact location using the global state.
	\begin{table}
		\vspace{16px}	
		\caption{Physical Parameter of the planar robot}
		\centering
		\begin{tabular}{c c c}
			\hline
			\vspace{0pt}
			Parameter & Unit & Value\\
			\hline
			$m$ & [kg] & 2.56\\
			$I_{\theta}$ & [kg$\cdot$m$^2$] & 0.04\\
			$L$  & [m]  & 0.3\\
			$l_{thigh}$ & [m] & 0.14\\
			$l_{shank}$ & [m] & 0.14\\
			$\tau_{max}$& [Nm]& 9.8\\
			\hline
		\end{tabular}
		\label{tab:phy_params}
	\end{table}
	It could be observed from (\ref{eq:dyn}) that the translational and rotational dynamics are linear in terms of the spatial wrench $\bm{\mathcal{F}}_s$. Let the spatial wrench trajectory be parametrized by the \Bezier\ polynomial with coefficient $\bm{\alpha}_{\mathcal{F}}=[\bm{\alpha}_{f_x}, \bm{\alpha}_{f_z}, \bm{\alpha}_{\tau_y}]^T$, then the trajectories $\dot{\bm{q}}(t),\bm{q}(t)$ are also parametrized by \Bezier\ polynomials with coefficients $\bm{\alpha}_{\dot{q}}$ and $\bm{\alpha}_q$, respectively. Given initial conditions $\dot{\bm{q}}_0,\bm{q}_0$, coefficients $\bm{\alpha}_{\dot{q}}$ and $\bm{\alpha}_q$ could be obtained by linear operations.
	\begin{equation}
	\bm{\alpha}_{\dot{q}}=\mathcal{L}(\bm{\alpha}_{\mathcal{F}},\dot{\bm{q}}_0),~\bm{\alpha}_{q}=\mathcal{L}(\bm{\alpha}_{\dot{q}},\bm{q}_0)
	\end{equation}
	where the linear operation $\mathcal{L}(\cdot)$ is defined in Appendix \ref{app:int_Bezier_poly}.
	
	Although the dynamics in (\ref{eq:dyn}) are linear in terms of the spatial wrench, the coupling between forces $\bm{f}_i$ and the moment created by the force $\tau_i^y$ imposes bilinear constraints on the feasible wrench. Section \ref{sec:FWP} presents how these constraints are represented using the feasible wrench polytope (FWP).
	
	The state and control trajectories during stance are polynomials parametrized by the \Bezier\ coefficients $\bm{\alpha}_{\mathcal{F}}$. The state trajectory during aerial phase could be parametrized by the second-order polynomial. To make the kinodynamic motion planning problem finite dimensional, the continuous time trajectory $\bm{\mathcal{F}}_s(t),\dot{\bm{q}}(t),\bm{q}(t)$ are sampled at a time sequence $\{t_k | k=1,2,\cdots, N_t\}$, where $N_t$ is the number of nodes during stance phase.

	\subsection{C-space Discretization}\label{sec:C-space}
	
	The configuration space (C-space) $\bm{\Omega}\subset\mathbb{R}^3$ of the robot is the set of configurations that the robot could reach during stance phase without violating kinematic constraints. The set of constraints that define the C-space is 
	\begin{subequations}\label{eq:C_space}
		\begin{align}
		\bm{\Omega}:=\{\bm{q}\in\mathbb{R}^3 |
		& 0\leq \hat{\bm{n}}^T\cdot(\bm{p}^h_{i}-\bm{p}^c_{i})\\
		& 0\leq \hat{\bm{n}}^T\cdot(\bm{p}^k_{i}-\bm{p}^c_{i})\\
		& r_{min}\leq |\bm{p}^h_{i}-\bm{p}^c_{i}|_2 \leq r_{max}\\
		& \bm{q}_{min} \leq \bm{q} \leq \bm{q}_{max} \},
		\end{align}
	\end{subequations}
	where $\hat{\bm{n}}$ is the normal vector of the terrain; $\bm{p}^h, \bm{p}^k$ are the hip and knee joint positions of the leg $i\in \{\textbf{\textit{b}}ack,\textbf{\textit{f}}ront\}$ and $\bm{p}^c$ is the foot contact position. (\ref{eq:C_space}a) and (\ref{eq:C_space}b) prohibit the hip and knee from penetrating the terrain surface; (\ref{eq:C_space}c) sets boundaries $r_{min}, r_{max}$ on the leg extension, using two-norm $|\cdot|_2$; (\ref{eq:C_space}d) is the box constraint on $\bm{q}$. The C-space is calculated based on the following assumption.
	\begin{assumption}[Constant Stance Width]\label{as:stance_width}
		The distance between two contact feet is equal to the body length $L$.
	\end{assumption}

\begin{figure}
		\centering
		\resizebox{0.95\linewidth}{!}{\includegraphics{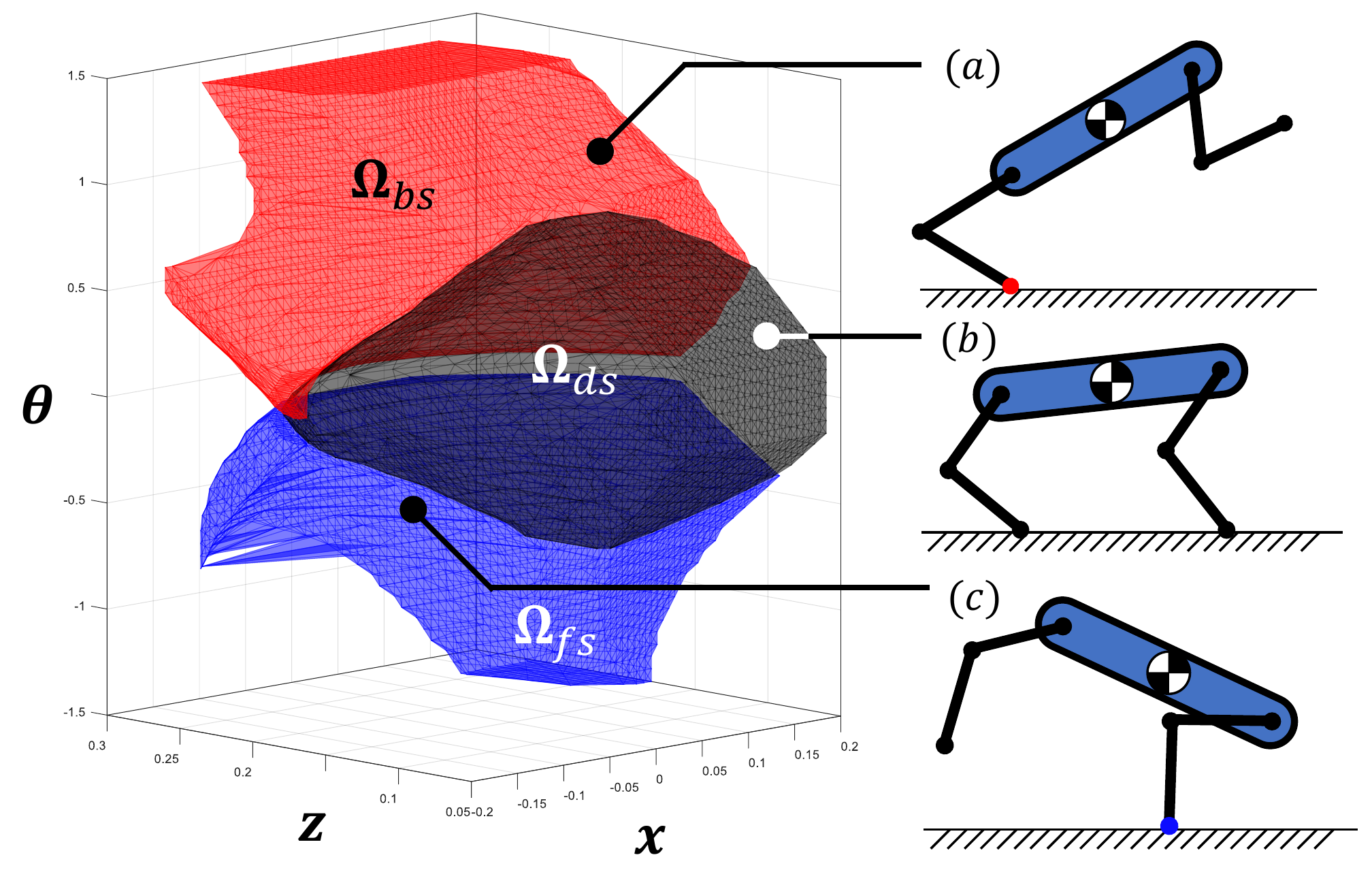}}
		\caption{(left) The 3-dimensional (3D) configuration space of the robot; (right) The illustration of robot configuration in each stance mode (a) back stance (b) double stance (c) front stance.}
		\label{fig:ws_and_contact}
	\end{figure}

	This assumption removes the dependency on the stance width, simplifying the C-space construction. Since all quantities in (\ref{eq:C_space}) could be retrieved through kinematic calculation given the robot configuration $\bm{q}$, the C-space could be clearly defined once the robot parameters are given. Fig. \ref{fig:ws_and_contact} shows the C-space of the robot with parameters in Table \ref{tab:phy_params}.
	
	During the stance phase, the robot could be in one of three stance modes, namely, front, double and back stance. When the robot could take double stance, it could also take either front or back stance by lifting one of its legs, complicating the contact scenario. To simplify the choice of stance mode, the following assumption is made.
	
	\begin{assumption}[Preference on Double Stance]\label{as:double_stance}
		The robot prefers double stance to single stance, and would be in double stance whenever possible.
	\end{assumption}

	Based on the observation that double stance provides more control authority compared with single stance, assumption \ref{as:double_stance} establishes a one-to-one mapping between robot configuration $\bm{q}$ and stance mode. As shown in Fig. \ref{fig:ws_and_contact}, the C-space is divided into three disjoint regions corresponding to front stance $\bm{\Omega}_{fs}$ (blue), double stance $\bm{\Omega}_{ds}$ (black), and back stance $\bm{\Omega}_{bs}$ (red), or equivalently,
	\begin{equation}\label{eq:3C-space}
	\bm{\Omega}:=\bm{\Omega}_{bs}\cup \bm{\Omega}_{fs} \cup \bm{\Omega}_{ds}.
	\end{equation}
	An illustration corresponding to each stance mode is visualized on the right panel of Fig. \ref{fig:ws_and_contact}. Due to the one-to-one mapping between robot configuration $\bm{q}$ and stance mode, the gait sequence planning is encoded into the C-space construction. Hence, a state trajectory $\bm{q}(t)$ in $\bm{\Omega}$ also contains the gait sequence information.

	As shown in Fig. \ref{fig:ws_and_contact}, the C-space is a non-convex set. To tackle this problem, the C-space $\bm{\Omega}$ is discretized into $N_d\in\mathbb{Z}^+$ convex polytopic cells, denoted by $\bm{c}_{i}\subset \bm{\Omega},i\in\{1,\cdots,N_d\}$, where $N_d$ is the total number of cells. The union of the cells is contained within the C-space, namely, $\cup_{i=1}^{N_d} \bm{c}_{i}\subset \bm{\Omega}$. For simplicity, tetrahedrons are used to discretize the C-space, whose distribution is designed such that each cell resides within the same stance region. The geometry of each cell is encoded by linear inequalities
	\begin{equation}
	\bm{A}_{i}^{geo}\cdot\bm{q}\leq \bm{b}_{i}^{geo}, i=1,\cdots,N_d,
	\end{equation}
	where the matrices $\bm{A}_{i}^{geo}$ and $\bm{b}_{i}^{geo}$ delineate the cell $\bm{c}_i$ using the half-plane representation ($\mathcal{H}$-Rep).
	
	\subsection{Feasible Wrench Polytope}\label{sec:FWP}
	This section presents a formulation of the feasible wrench polytope ($FWP$) that is pertinent to the robot system studied in this work. For a more comprehensive derivation of the $FWP$, please refer to \cite{Romeo_18}.
	\begin{figure}
		\centering
		\resizebox{1.0\linewidth}{!}{\includegraphics{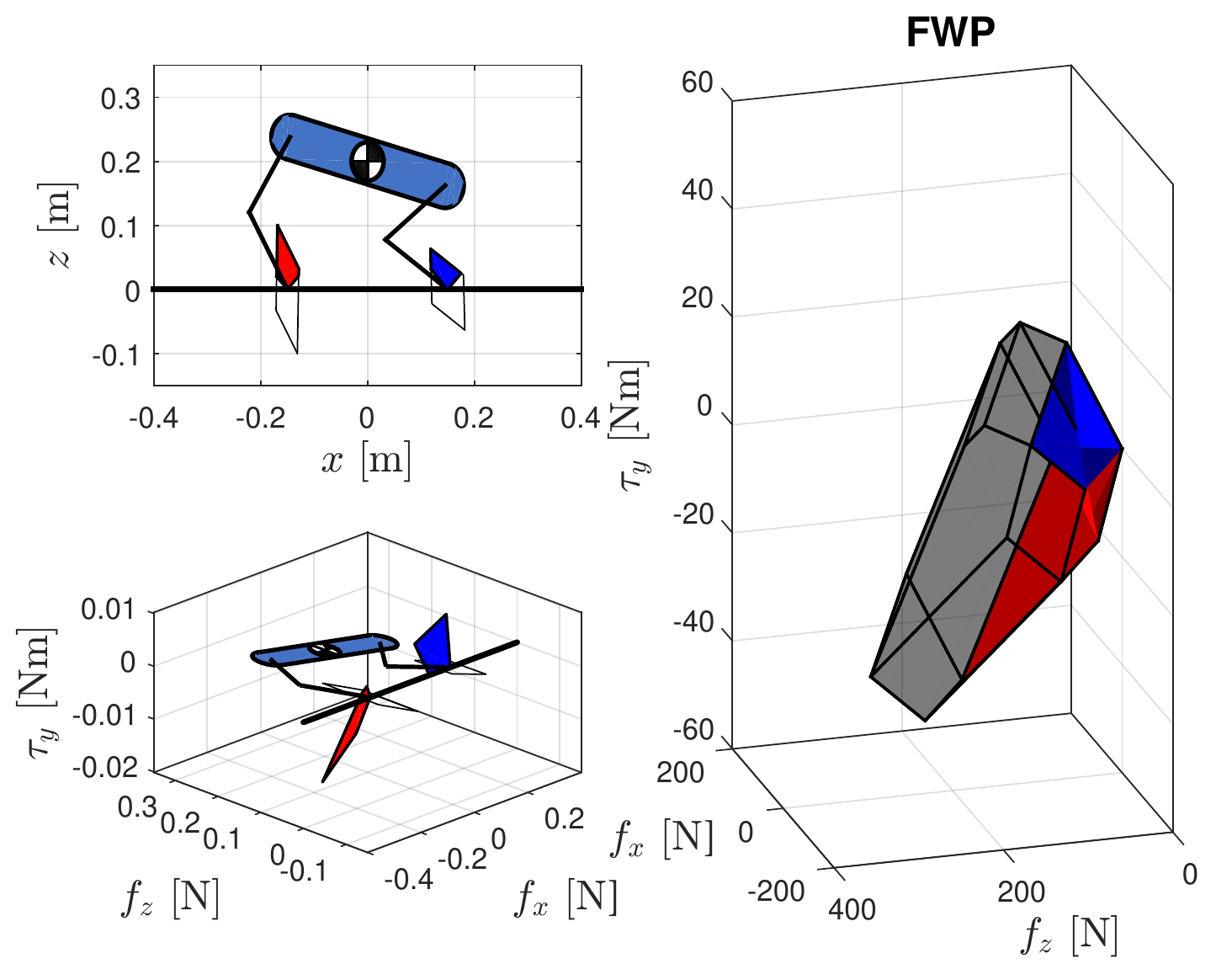}}
		\put(-210,170){(a)}
		\put(-210,60){(b)}
		\put(-95,150){(c)}
		\caption{The FWP of the two-legged robot. (a) The $FFP$ of front (blue) and back (red) legs. (b) The $FWP$ of both legs plotted in the 3D wrench space. (c) The $FWP$ of the robot is the Minkowski sum of the $FWP$ of both legs.}
		\label{fig:FWP}
	\end{figure}
	
	\subsubsection{\textbf{FWP of One Leg}}
	First we define the feasible force polytope ($FFP$) for one leg as
	\begin{subequations}\label{eq:FFP}
	\begin{align}
	FFP:=\{\bm{f}\in\mathbb{R}^2~|~
	& |\bm{J}^T \bm{f}|_{\infty}\leq \tau_{max}\\
	& |\bm{f}-\hat{\bm{n}}^T\bm{f}|_{\infty} \leq \mu \hat{\bm{n}}^T\bm{f}\},
	\end{align}
	\end{subequations}
	where $\bm{J}$ is the Jacobian matrix; $\tau_{max}$ is the joint torque limit; $\mu$ is the coefficient of friction. Based on the assumption \ref{as:light_leg}, the GRF is mapped to joint torque via $\bm{\tau} = \bm{J}^T \bm{f}$. Furthermore, assumption \ref{as:stance_width} implies that given the robot configuration $\bm{q}$, joint angles could be calculated through inverse kinematics. The inequality (\ref{eq:FFP}a) encodes the joint torque constraint $|\bm{\tau}|_{\infty}<\tau_{max}$, where $|\cdot|_{\infty}$ is the infinity norm. The inequality  (\ref{eq:FFP}b) represents the friction cone constraint.
	
	The $FWP$ of one leg is the set of spatial wrenches that could be provided by the $FFP$
	\begin{equation}\label{eq:FWP_oneLeg}
	FWP_i:=\{\bm{\mathcal{F}}_i\in\mathbb{R}^3~|~\bm{\mathcal{F}}_i=\begin{bmatrix}
	\bm{f}^k_i\\
	\bm{r}_i\wedge \bm{f}^k_i
	\end{bmatrix},\bm{f}^k_i\in FFP_i\},
	\end{equation}
	where $\bm{f}^k_i$ is the $k^{th}$ vertex of $FFP$ for contact point $i$. Note that the $FWP$ is defined using the vertex-representation of a polytope ($\mathcal{V}$-Rep). Fig. \ref{fig:FWP}(b) shows an example of $FWP$s of front leg (blue) and back leg (red), which are 2-D polytopes embedded in the 3-D wrench space.
	
	\subsubsection{\textbf{FWP of Two Legs}}
	The $FWP$ when the two-legged robot is in double stance is defined as the Minkowsi sum \cite{varadhan2004accurate} of the $FWP$ created by both contact legs,
	\begin{equation}\label{eq:FWP_ds}
	FWP_q = \bigoplus_{i=1}^{2} FWP_i,
	\end{equation}
	where $FWP_q$ indicates the $FWP$ of the robot at configuration $\bm{q}$. The Minkowski sum of two sets $X$ and $Y$ is $X\oplus Y:=\{x+y|x\in X,y\in Y\}$. Fig. \ref{fig:FWP}(c) shows an example of $FWP_q$ as the Minkowski sum of the $FWP$ of both legs. Assumption \ref{as:double_stance} implies that when $\bm{q}\in \bm{\Omega}_{ds}$, the corresponding $FWP_q$ is defined as in (\ref{eq:FWP_ds}).
	
	For a cell $\bm{c}_i$ in the C-space discretization, its representative $FWP$ is defined as
	\begin{equation}\label{eq:FWP_cell}
	FWP_{c_i} = \begin{cases}
	\cap_{k=1}^{4} FWP_{c_i}^k, & \text{double stance}\\
	FWP_{c_i}^{cbsv}, & \text{single stance},
	\end{cases}
	\end{equation}
	where $FWP_{c_i}^k$ is the $FWP$ of the $k^{th}$ node of the cell $\bm{c}_i$. $FWP_{c_i}^{cbsv}$ is the $FWP$ at the Chebyshev center \cite{boyd2004convex}, which is the center of the largest Euclidean ball that lies in a polytope. The choice of $FWP$ for double stance in (\ref{eq:FWP_cell}) is a conservative approximation, since the wrench in $FWP_{c_i}$ could be achieved at all 4 vertices in the tetrahedron of the cell $\bm{c}_i$. This formulation provides robustness when the robot is in the double stance since it is an inner approximation. In comparison, for a cell in the single stance, the $FWP$ at each vertex degenerates to a 2-D polytope due to the coupling between forces and moment. Since each node corresponds to a different $\bm{r}_i$, the $FWP$s have no intersection except at the origin. Therefore, the $FWP$ for cells in the single stance is defined at the Chebyshev center of the cell.
	
	The $FWP_{c_i}$ could be represented using the half-plane representation ($\mathcal{H}$-Rep) consisting of a set of linear constraints
	\begin{equation}\label{eq:FWP}
	\bm{A}^{fwp}_i\cdot \bm{\mathcal{F}}_s \leq \bm{b}^{fwp}_i, i = 1,\cdots,N_d,
	\end{equation}
	where $\bm{A}^{fwp}_i$ and $\bm{b}^{fwp}_i$ are the matrices that describe half-planes. The single stance $FWP$ is subject to equality constraints $\bm{A}^{fwp}_{i,e}\cdot \bm{\mathcal{F}}_s = \bm{b}^{fwp}_{i,e}$, which could also be incorporated into the form of inequality constraint as in (\ref{eq:FWP}).
	
	Note that for a given physical parameter of the robot and a C-space discretization, $FWP_{c_i}$ only needs to be computed once.
	
	\subsection{Mixed-integer Wrench Constraint}
	The nonlinear and non-convex wrench constraint is imposed in a piecewise constant fashion over the discretized C-space, enabling a mixed-integer convex formulation. A binary matrix $\bm{B}^{cs}\in\{0,1\}^{N_t\times N_d}$ is constructed such that $\bm{B}^{cs}_{i,j}=1$ indicates that $\bm{q}(t_{i})$ is within cell $\bm{c}_j$ and the spatial wrench should be chosen within $FWP_{c_i}$
	\begin{equation}\label{eq:implies}
	\begin{aligned}
	\bm{B}^{cs}_{i,j} &\implies \bm{A}_{j}^{geo}\cdot\bm{q}(t_{i})\leq \bm{b}_{j}^{geo}\\
	&\implies \bm{A}^{fwp}_j\cdot \bm{\mathcal{F}}_s(t_i) \leq \bm{b}^{fwp}_j,
	\end{aligned}
	\end{equation}
	where the \textit{implies} operator ($\implies$) in (\ref{eq:implies}) is implemented using the big-M formulation \cite{schouwenaars2001mixed}. Additional constraints are imposed for physical feasibility
	\begin{equation}\label{eq:sum_bigM}
	\sum_{j=1}^{N_d} \bm{B}^{cs}_{i,j}=1, \; \forall i = 1,\cdots,N_t
	\end{equation}
	which requires that at each time step $t_{i}$, the robot state $\bm{q}(t_i)$ can only reside within one cell $\bm{c}_j$.
	
	\subsection{Aerial Phase Constraints}\label{sec:aerial}
	The objective of the kinodynamic motion planner is to reach the goal region through a series of jumping motions, which involves both stance and aerial phases. During the aerial phase, the robot is airborne and only subject to gravity, whose trajectory is described by
	
	\begin{equation}\label{eq:aerial}
		\begin{bmatrix}
		\bar{\bm{q}}\\
		\dot{\bm{q}}
		\end{bmatrix}_{i+1}^{TD}
		=
		\begin{bmatrix}
		\bar{\bm{q}}\\
		\dot{\bm{q}}
		\end{bmatrix}_i^{TO}
		+
		\begin{bmatrix}
		\dot{\bm{q}}_i^{TO}\\
		\bm{a}_g
		\end{bmatrix}T_{i,air}
		+
		\begin{bmatrix}
		\frac{1}{2}\bm{a}_g\\
		\bm{0}
		\end{bmatrix}T_{i,air}^2
	\end{equation}
	where the superscripts $(\cdot)^{TO}$ and $(\cdot)^{TD}$ indicate variables at take-off and touch-down, respectively; $T_{i,air}$ is the aerial time of the $i^{th}$ jump; where $i=1,\cdots N_{jp}$ and $N_{jp}$ is the number of jumps. The state $\bar{\bm{q}}_i^{TD}$ is the sum of the global state $\bm{q}_i^{TD}$ and the local state at the start of the next stance phase $\bm{q}_i^{start}$. Similarly, $\bar{\bm{q}}_i^{TO} = \bm{q}^{TD}_i + \bm{q}_i^{end}$, where $\bm{q}_i^{end}$ is the end state of the $i^{th}$ stance phase. This dichotomy of global and local state is convenient for imposing wrench constraint (\ref{eq:implies}) on the local state and choosing contact location using the global state.
	
	Although the stance time $T_{st}$ of each jump is set to a constant to simplify the problem, the aerial time $T_{air}$ is an optimization variable. Hence, the bilinear terms $\dot{\bm{q}}_i^{TO} T_{i,air}$ in (\ref{eq:aerial}) leads to non-convex constraints. To this end, methods such as using the McCormick Envelope \cite{mccormick1976computability} to approximate the bilinear terms have been applied on planning aggressive motions of legged robots \cite{valenzuela2016mixed}. McCormick Envelope is used in our formulation because it could provide a close approximation with a relatively small number of binary variables since the range of each quantity could be empirically bounded from simulation and experiment. Using a similar technique, the quadratic term $T_{i,air}^2$ in (\ref{eq:aerial}) is approximated by a piecewise affine function.

	\subsection{Foothold Position}\label{sec:contact_location}
	With assumption \ref{as:stance_width}, the foothold position choice is simplified to finding $\bm{q}^{TD}$ on the given terrain. To simplify terrain geometry, the following assumption is made.
	\begin{assumption}[Terrain]\label{as:terrain}
		The terrain consists of segments which are modeled by piecewise affine functions. 
	\end{assumption}

	A binary matrix $\bm{B}^{fp}\in \{0,1\}^{N_{s}\times N_{j}}$ is constructed to assign foothold positions, where $N_{s}$ is the total number of terrain segments. $\bm{B}^{fp}_{i,j}=1$ implies that at the $j^{th}$ jump, the $\bm{q}^{TD}_j$ lies on the $i^{th}$ terrain segment $\bm{seg}_i$
	\begin{subequations}\label{eq:foot_constraint}
		\begin{align}
		&\bm{B}^{fp}_{i,j}=1 \implies \bm{q}^{TD}_j \in \bm{Q}_i\\
		&\sum_{i=1}^{N_{s}}\bm{B}^{fp}_{i,j}=1, \; \forall j =1,\cdots,N_{jp},
		\end{align}
	\end{subequations}
	where $\bm{Q}_i$ refers to the $i^{th}$ terrain segment. At the end of the last jump, the robot should reach the goal region
	\begin{equation}\label{eq:goal}
	\bm{q}_{N_{jp}}^{TD}\in \bm{Q}_{goal}
	\end{equation}
	where $\bm{q}_{N_{jp}}^{TD}$ is the touchdown state after the final jump and $\bm{Q}_{goal}$ is the goal region.

	\subsection{The Mixed-integer Convex Program}
	The kinodynamic motion planning problem of a planar two-legged robot could be transcribed to a MICP. The decision variable vector for this particular problem is $\bm{x}_{opt}=[\bm{\alpha}_{\mathcal{F}},\bm{q}_0,\dot{\bm{q}}_0,\bm{q}^{TD},\bm{T}_{air},\bm{B}^{cs},\bm{B}^{fp}]^T$. The complete formulation of the MICP is:
	\begin{subequations}\label{eq:MICP}
		\begin{align}
		\underset{\bm{x}_{opt}}{\text{min.}}~& f(\bm{x}_{opt})\\
		\text{s.t.}~& \bm{q}(t_i)\in\bm{c}_k \subset\bm{\Omega}\\
		& \ddot{\bm{q}}(t_i)=\bm{D}^{-1}\bm{\mathcal{F}}_s(t_i) + \bm{a}_g\\
		& \bm{\mathcal{F}}_s(t_i)\in FWP_{c_k}\\
		& \bm{q}_j^{TD}\in\bm{Q}_{l}\\
		& \bm{q}_0 \in \bm{Q}_{init},~\bm{q}_{N_{jp}}^{TD} \in \bm{Q}_{goal}\\
		& \text{aerial phase constraint: } (\ref{eq:aerial})\\
		& i=1,\cdots,N_t;  j=1,\cdots,N_{jp}\\
		& k=1,\cdots,N_d; l=1,\cdots,N_s
		\end{align}
	\end{subequations}
	where (\ref{eq:MICP}b) is the kinematic constraint baked into the C-space; (\ref{eq:MICP}c) is the dynamic constraint and (\ref{eq:MICP}d) is the wrench constraint. (\ref{eq:MICP}e) constrains the touchdown state to be on the terrain; (\ref{eq:MICP}f) are the boundary condition constraints, where $\bm{Q}_{init}$ is the feasible set for the initial condition. $f(\bm{x}_{opt})$ is a task-specific convex objective function. For example, $f(\cdot)$ could be the deviation from the goal  $|\bm{q}^{TD}_{N_{jp}}-\bm{q}_{goal}|_2$, which makes the problem a mixed-integer quadratic program (MIQP); or it could be $-x^{TD}_{N_{jp}}$ to maximize the horizontal jumping distance, which leads to a mixed-integer linear program (MILP); $f(\cdot)$ could also be set as a constant value to solve a feasibility problem.
	
	The MICP problem is formulated in MATLAB using YALMIP \cite{Lofberg2004}. The computational geometry calculation related to $FWP$ is done using the Multi-Parametric Toolbox 3 (MPT3) \cite{MPT3}. The MICP is solved by the solver Gurobi \cite{gurobi}. All of the computation is performed on a desktop with 2.9 GHz Intel i7.
	
	
	\section{Results}
	\label{sec:results}

	To validate the proposed kinodynamic motion planning algorithm, jumping experiments are conducted on the robot. Experiment results for both jumping forward and backward, together with the simulation result of a dynamic Parkour motion are presented. Note that the trajectories of all three motions are solved by the proposed MICP without any initial guesses.
	
	\subsection{Experimental Setup}
	
	As shown in Fig. \ref{fig:jumpUp}, the planner two-legged robot similar to the one used in \cite{li2020centroidal} is composed of the torso made of a carbon fiber tube and two legs modules, which enable dynamic maneuvers that demands high joint torques \cite{Ding_Design}. Each joint of the robot is equipped with a RLS-RMB20 magnetic encoder, and an inertial measurement unit (IMU) is mounted for state estimation, which is the same as that in \cite{Ding_19_MPC}. The robot is mounted on the end of a passive boom system with a radius $R_{boom}=1.25$ m. Two encoders are installed at the base of the boom to measure the global positioning of the robot, and another encoder is mounted at the connection between the tip of the boom and the robot to measure the pitch angle $\theta$. The robot is externally powered and the Elmo Gold Twitter amplifiers are mounted on the boom. The control loop runs at 4kHz on an Intel i5 desktop in Simulink Real-Time. The physical parameters of the robot could be found in Table \ref{tab:phy_params}.

	\subsection{Sagittal Plane Control}\label{sec:sagittal}
	The spatial wrench trajectory obtained from solving the MICP is distributed to the GRF in the sagittal plane using the closed-chain-constrained operational-space control \cite{Hutter_14_IJRR}. Similar to the frontal plane controller in \cite{haewon_IJRR}, a linear operator $\bm{\Pi}\in\mathbb{R}^{4\times 3}$ is calculated to map the spatial wrench $\bm{\mathcal{F}}_s$ to the joint torque $\bm{u}_{sag} = \bm{\Pi} ~\bm{\mathcal{F}}_s$. Let $\bm{q}_{sag}:=[\bm{q}^T,\bm{q}_l^T]^T \in\mathbb{R}^{7\times 1}$ be the generalized coordinates of the sagittal dynamics, where $\bm{q}_l$ is the vector of joint angle. The no-slip ground contact constraints are
	\begin{equation}\label{eq:foot_cst}
	\begin{aligned}
	&\dot{\bm{p}}_{foot}=\bm{J}_{foot}\dot{\bm{q}}_{sag}=0\\
	&\ddot{\bm{p}}_{foot}=\bm{J}_{foot}\ddot{\bm{q}}_{sag} + \dot{\bm{J}}_{foot}\dot{\bm{q}}_{sag} = 0,\\
	\end{aligned}
	\end{equation}
	where $\bm{p}_{foot}$ is the foot position, and the foot Jacobian $\bm{J}_{foot}\in\mathbb{R}^{4\times 7}$ is partitioned as $\bm{J}_{foot}=[\bm{J}_B, \bm{J}_l]$. The dynamics of the sagittal plane system are
	\begin{equation}
	\begin{bmatrix}
	\bm{D} & 0 &-\bm{J}_B^T\\
	0 & 0 & -\bm{J}_l^T\\
	\bm{J}_B & \bm{J}_l & 0
	\end{bmatrix}
	\begin{bmatrix}
	\ddot{\bm{q}}\\
	\ddot{\bm{q}}_l\\
	\bm{\lambda}
	\end{bmatrix}=
	\begin{bmatrix}
	0\\
	\bm{u}_{sag}\\
	-\dot{\bm{J}}_{foot}\dot{\bm{q}}_{sag}
	\end{bmatrix}
	\end{equation}
	where $\bm{\lambda}$ is the Lagrange multiplier associated with the constraints in (\ref{eq:foot_cst}). By solving the system of equations one could derive $-\bm{J}_B^T\bm{J}_l^{-T}\bm{u}_{sag} = \bm{D}\ddot{\bm{q}} = \bm{\mathcal{F}}_s.$ Thus, as one of many possible solutions, $\bm{\Pi}$ is selected as $\bm{\Pi}=-(\bm{J}_B\bm{J}_l^{-T})^{\dagger}$, where $(\cdot)^{\dagger}$ provides the Moore-Penrose pseudoinverse.
	
	Using the saggittal plan control, the feedforward wrench trajectory $\bm{\mathcal{F}}_{ff}$ from MICP and a proportional-derivative (PD) feedback controller is used to track the desired state trajectory. $\bm{\mathcal{F}}_{fb} = \bm{K}_p (\bm{q}_d - \bm{q}) + \bm{K}_d (\dot{\bm{q}}_d - \dot{\bm{q}})$, where $\bm{K}_p,\bm{K}_d$ are diagonal gain matrices; $\bm{q}_d$ and $\dot{\bm{q}}_d$ are the desired state trajectories. During the swing phase, a workspace PD controller is applied on the swing foot to track the prescribed swing trajectory.
	
	\subsection{Jump On Platforms}
	\begin{figure}
		\centering
		\resizebox{1\linewidth}{!}{\includegraphics{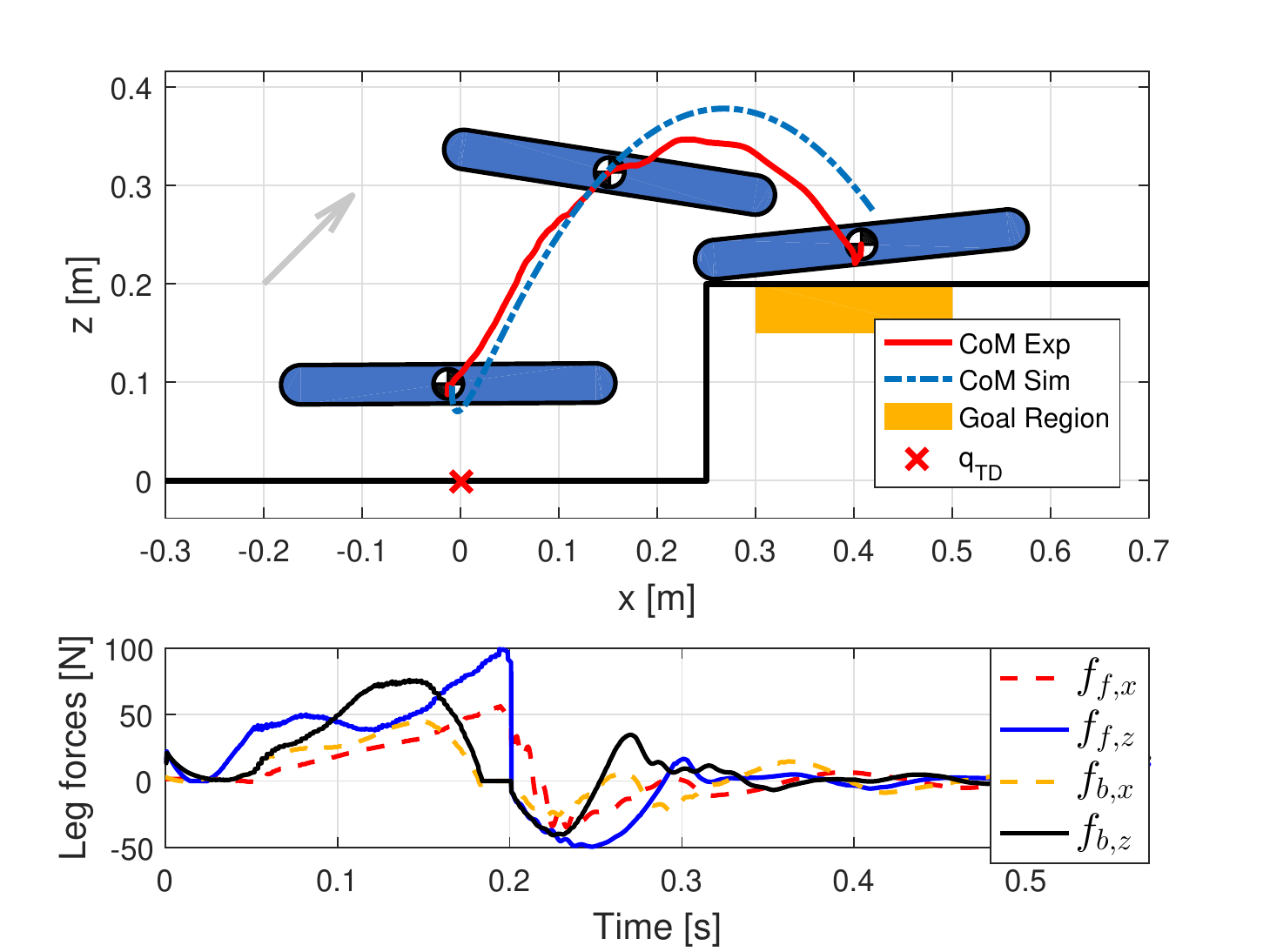}}
		\put(-210,160){(a)}
		\put(-210,50){(b)}
		\caption{The result of the experiment where the robot jumped forward on a platform. (a) CoM trajectories of simulation (blue) and experiment (red); the orange area is the goal region; the gray arrow indicates the jumping direction. (b) The leg forces during the stance phase $t\leq 0.2~s$ and the aerial phase.}
		\label{fig:exp_jumpUp_forward}
	\end{figure}
\begin{figure*}
		\centering
		\resizebox{0.95\linewidth}{!}{\includegraphics{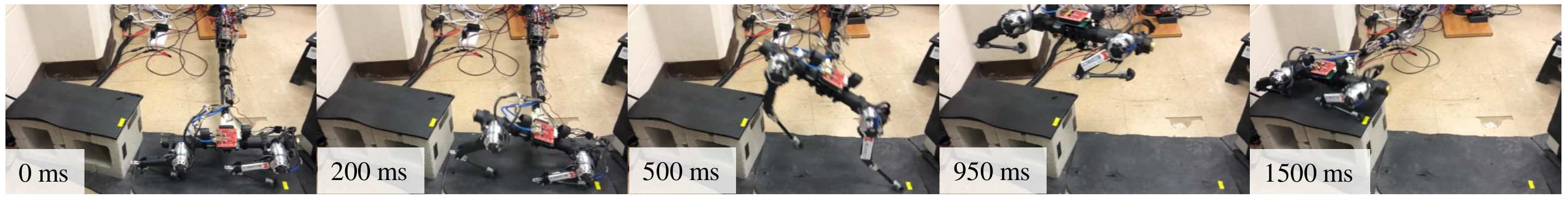}}
		\caption{Sequential snapshots of the experiment where the robot jumps backwards onto a 0.2 m high platform to reach the goal region.}
		\label{fig:jumpUp_backward}
	\end{figure*}

	As shown in Fig. \ref{fig:jumpUp}, the robot jumps forward on a 0.2 m high platform. Both simulation and experimental results are shown in Fig. \ref{fig:exp_jumpUp_forward}, where the CoM trajectory from the simulation is shown in blue and that from the experiment is shown in red. The leg forces are shown in Fig. \ref{fig:exp_jumpUp_forward} (b) during the stance phase ($t\leq0.2$ s) and the aerial phase. Another experiment where the robot jumps back onto the platform is shown in Fig. \ref{fig:jumpUp_backward}, where the sequential snapshots show that the solution involves using large body pitch oscillation to aid the robot to jump on the platform.
	
	For these less dynamic jumping motions, only double stance region $\bm{\Omega}_{ds}$ is used. The number of variables for both motions is 216 (27 continuous, 189 integer), and the computational time to solve the MICP is 0.84 s for the jumping forward problem and 5.94 s for the jumping backward problem. The solve time difference may be explained by that the knee-bending-back configuration provides more forward force authority. Additionally, the knee-bending-forward configuration imposes stricter collision avoidance constraints between the knee and terrain. The \Bezier\ coefficients of the wrench trajectories are summarized in Table \ref{tab:B-co}.
	
	\begin{table}
		\centering
		\begin{tabular}{c c c}
			\hline
			Experiment & Wrench & \Bezier~ Coefficients\\
			\hline
			& $\bm{\alpha}_{f_x}$ & 0.0, -7.6, 33.8, -33.7, 90.1, 0.0\\
			Jump forward 	& $\bm{\alpha}_{f_z}$ & 25.1, -69.0, 152.0, -50.0, 262.7, 0.0\\
			& $\bm{\alpha}_{\tau_y}$ & 0.0, 5.5, -25.3, 25.5, -5.8, 0.0\\
			\hline
			& $\bm{\alpha}_{f_x}$ & 0.0, 198.4, -404.8, 296.7, -135.5, 0.0\\
			Jump backward 	& $\bm{\alpha}_{f_z}$ & 25.1, -171.9, 631.8, -786.8, 623.5, 0.0\\
			& $\bm{\alpha}_{\tau_y}$ & 0.0, 44.5, -79.4, 48.2, -14.5, 0.0\\
			\hline
		\end{tabular}
		\caption{\Bezier~coefficient for the jumping on platform experiments}
		\label{tab:B-co}
	\end{table}
	
	\subsection{Parkour Motion}\label{sec:parkour}
\begin{figure}
		\centering
		\resizebox{1\linewidth}{!}{\includegraphics{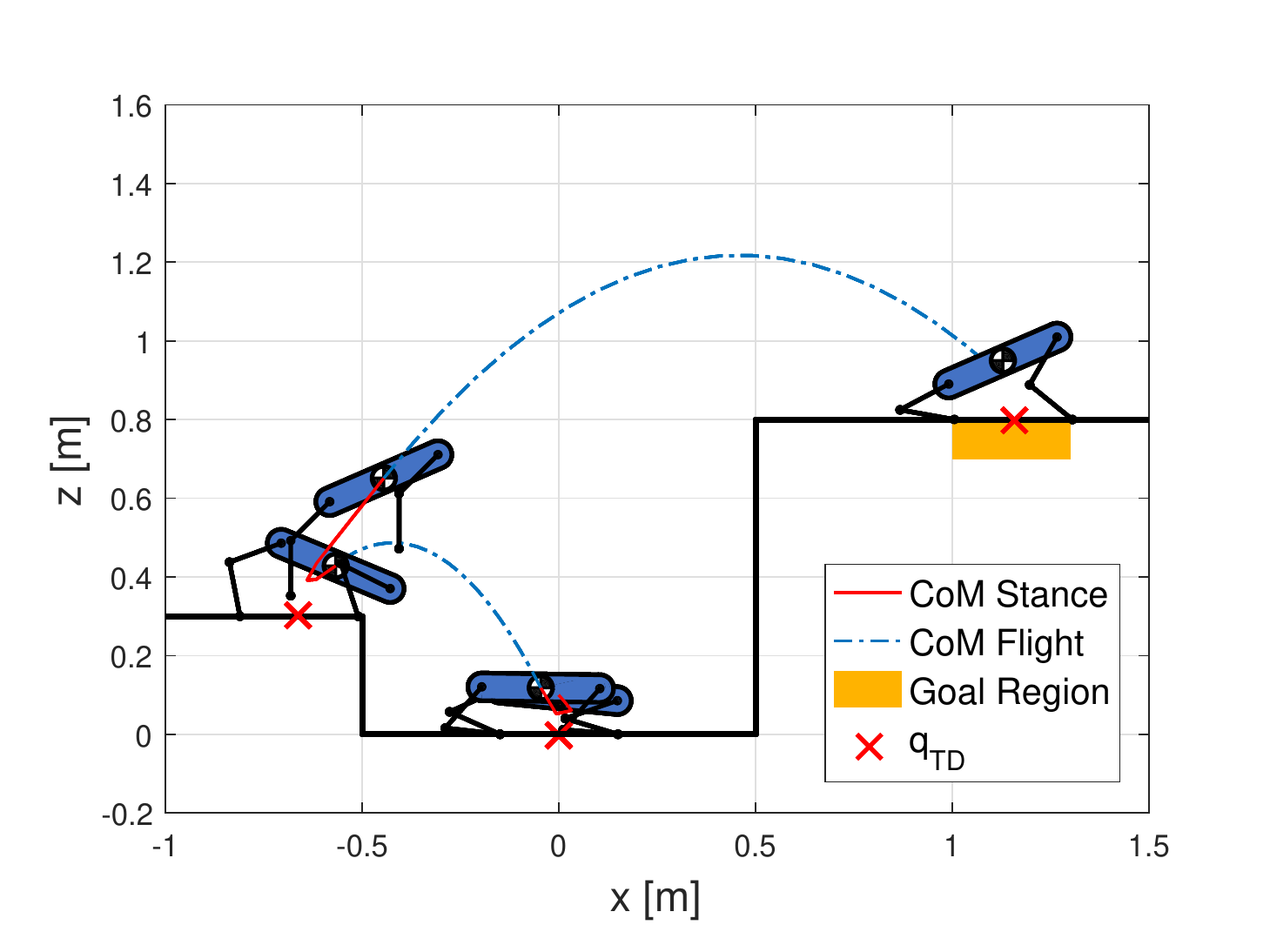}}
		\caption{Simulation result of the Parkour motion. The proposed formulation can find the strategy of utilizing the left platform as a stepping-stone to reach the goal region on the high platform.}
		\label{fig:sim_Parkour}
	\end{figure}

	The proposed kinodynamic motion planning framework could generate plans that traverse terrains that require complex maneuvers. For example, Fig. \ref{fig:sim_Parkour} shows one problem setup where the goal region is on the high platform, and the robot cannot reach it with a single jump due to actuation limitations. As shown in Fig. \ref{fig:sim_Parkour}, the MICP provides the solution where the left platform is used as a stepping stone towards the goal region. By making two jumps, the proposed framework solves the problem without initial guess nor user input about the step planning. With the grid resolution $N_{bs} = 10,N_{fs} = 10,N_{ds} = 21$, the MICP involves 485 variables (53 continuous, 432 integer), and the computational time is 27 s. The algorith
	m utilized double stance and back stance. The back stance is used towards the end of the second jump, presumably to take advantage of the extra kinematic reachability in body pitch.
	
	This simulation result showcases one of the advantages of mixed-integer program based motion planning algorithms, which is that it could reason about making discrete decisions.
	
	\section{Discussion}\label{sec:discusstion}
	In this section, we discuss some of the findings and limitations of this work. This work has extended the single-leg model in  \cite{Ding_18} to the multi-legged model, which poses unique challenges since different stance modes dictate whether the system is over-actuated or under-actuated. 
	
	Compared with \cite{BAC}, we focus on generating dynamic jumping motions that are not limited to the vicinity of the nominal pose. This work is most similar to \cite{valenzuela2016mixed} in spirit, where dynamic legged motions are generated via MIQP. The construct of this work allows the joint torque constraint to be explicitly imposed using the notion of $FWP$. However, with the series of assumptions made intending to reduce the number of decision variables, the proposed method is only applicable to planner models. The stance time and number of jumps should also be chosen \textit{a priori}.
	
	In experiments, the sagittal plane control works well for the double stance. Nevertheless, for Parkour motion in Section \ref{sec:parkour}, the jump on the lower platform often involves undesirable single stance phase which the reactive sagittal plane control could not stabilize well. In hindsight, model-predictive control (MPC) may provide more robust performance since it could reason about the robot dynamics within the prediction horizon.

	\section{Conclusion and Future Work}\label{sec:conclusion}
	This paper proposed a novel MICP-based kinodynamic motion planning framework for aggressive jumping motions on 2-D multi-legged robots. The proposed method could produce highly dynamic jumping motions by solving for the centroidal motion, contact location, contact wrench and gait sequence simultaneously in a single MICP, with a global optimality certificate. The MICP approximates the nonlinear and non-convex constraints into piecewise convex constraints, and a preferred ordering on the robot stance modes is introduced to encode the gait sequence into the robot state trajectory. Both simulation and experiment show that the proposed planner could generate dynamically feasible motions on complex terrain. In the future, we plan to combine the MICP planner with MPC for robust jumping motions in experiments. We also envision combining sampling-based methods to tackle more complex terrains.
	
	\section{Acknowledgment}
	The authors would like to thank Prof. Kris Hauser, Prof. Jo$\tilde{\text{a}}$o Ramos and Dr. Zherong Pan for the helpful insightful discussions.
	
		\bibliographystyle{IEEEtran}	
		\bibliography{IROS20}

	\begin{appendices}
		\section{Integration of \Bezier\ Polynomial}\label{app:int_Bezier_poly}
		A \Bezier\ polynomial is a linear combination of a Bernstein polynomial basis \cite{dicsibuyuk2007generalization}, so the integration of a \Bezier\ polynomial is a linear operation \cite{Doha2011} on the \Bezier\ coefficients. For example, the linear relationship between wrench \Bezier\ coefficients and twist \Bezier\ coefficients is
		\begin{equation}\label{eq:bernstein_integration}
		\frac{M+1}{T_{st}} \bm{\Phi}(M,T_{st})\bm{\alpha}_{\dot{q}}=[\bm{D}^{-1}\bm{\alpha}_{\mathcal{F}}^T+\bm{a}_g,\dot{\bm{q}}_0]^T,
		\end{equation}
		where $M$ is the order of \Bezier\ polynomial; $T_{st}$ is stance duration; $\dot{\bm{q}}_0\in \mathbb{R}^{3}$ is the initial body twist; $\bm{\alpha}_{\dot{q}}\in \mathbb{R}^{(M+2)\times{3}}$ is the \Bezier\ coefficients for the spatial twist trajectory; $\bm{\Phi}\in \mathbb{R}^{(M+2)\times(M+2)}$ is a matrix whose elements are defined as
		\begin{equation}
		\bm{\Phi}_{i,j}:=
		\begin{cases}
		-1,& j=i=1,2,\cdots,M+1\\
		1,& j=i+1=2,3,\cdots,M+2\\
		\frac{T_{st}}{M+1},& i=M+2, j=1\\
		0,& \mbox{otherwise}.
		\end{cases}
		\end{equation}
		The linear operation $\bm{\alpha}_{\dot{q}}=\mathcal{L}(\bm{\alpha}_{\mathcal{F}},\dot{\bm{q}}_0)$ is obtained by inverting the matrix in front of $\bm{\alpha}_{\dot{q}}$ in (\ref{eq:bernstein_integration}). Similarly, the \Bezier\ coefficients of the configuration trajectory $\bm{q}(t)$ could also be integrated given initial configuration $\bm{q}_0 \in\mathbb{R}^3$. 
	\end{appendices}
	
\end{document}